\documentclass[journal,twoside,web]{ieeecolor}
\usepackage{generic}
\usepackage{cite}
\usepackage{amsmath,amssymb,amsfonts}
\usepackage{algorithmic}
\usepackage{graphicx}
\usepackage{algorithm,algorithmic}
\usepackage{hyperref}
\hypersetup{hidelinks}
\usepackage{textcomp}
\usepackage{multirow}
 \usepackage{booktabs}
\def\BibTeX{{\rm B\kern-.05em{\sc i\kern-.025em b}\kern-.08em
    T\kern-.1667em\lower.7ex\hbox{E}\kern-.125emX}}
\providecommand{\refname}{References}
\begin{document}
\title{Efficient Multi-Granularity Knowledge Transfer  for Radiology Report Generation}
\author{Xubin Zhong, Zheyu Zhang, Wenjian Qin, and 
Ning Wen
\thanks{This work was supported by the National Natural Science Foundation of China (82472065, W2432049), 
the Open Project Program of National Research Center for Translational Medicine at Shanghai (Grant No. NRCTM(SH)-2025-11), Ruijin Hospital, Shanghai Jiao Tong University School of Medicine, Shanghai, China and the Shanghai Explorer Program (24TS1414900).}
\thanks{Xubin Zhong is with the Center Rearch Institute of United Imaging, Shenzhen 518000, China. (e-mail: xubin.zhong@cri-united-imaging.com) }
\thanks{Zheyu Zhang is with the Global College, Shanghai Jiao Tong University, Shanghai 200240, China. (e-mail: jimzhang826@sjtu.edu.cn) }
\thanks{Wenjian Qin is the corresponding author. He is with the Shenzhen Institutes of Advanced Technology, Chinese Academy of Sciences, Shenzhen 518000, China. (e-mail: wj.qin@siat.ac.cn).}
\thanks{Ning Wen is the corresponding author. He is with the Department of Radiology \& Institute for Medical Imaging Technology, Ruijin Hospital, Shanghai Jiao Tong University School of Medicine, 197 Ruijin 2nd Road, Huangpu District, Shanghai, China. (e-mail: wn12479@rjh.com.cn).}
}

\maketitle

\begin{abstract}
Radiology report generation can automatically generate clinical descriptions from X-ray images, thereby significantly improving the efficiency of radiologists. 
This task is challenging because it requires medical knowledge to accurately identify diseases and describe them in a professional manner. However, existing methods often overlook the importance of enhancing medical knowledge in describing pivotal areas, a capability that requires models to effectively extract and aggregate knowledge at multiple levels of granularity. Accordingly, we herein propose a novel and compact Efficient Multi-Granularity Knowledge Transfer (\textbf{EMGKT}) method to address the above issues. First, we encode global knowledge embeddings using a medical vision-language  model, which provides contextual medical knowledge. Moreover, we devise a novel Fine-Grained Knowledge Distillation (FGKD) training task which efficiently extract fine-grained knowledge. Specifically, the FGKD training task contains teacher embeddings and student embeddings. Teacher embeddings are encoded using extra priors; while student embeddings are learned from the teacher embeddings through knowledge distillation. During inference, the student embeddings are used to enhance fine-grained knowledge while the teacher embeddings are discarded, resulting in negligible computational costs and no need for extra priors. 
Finally, we further develop a mixture of disease diagnosis expert  classifiers to enhance knowledge extraction. The classifiers are initialized using disease  embeddings and are modeled as different experts to address various granularity features. Notably, \textbf{EMGKT} can be efficiently applied  to most existing methods. Extensive  experiments are conducted on two widely-used public datasets and various baselines, which demonstrates the effectiveness and transferability of \textbf{EMGKT}.
\end{abstract}


\section{Introduction}

\begin{figure*}[t]
\centering
\includegraphics[width=1.\textwidth]{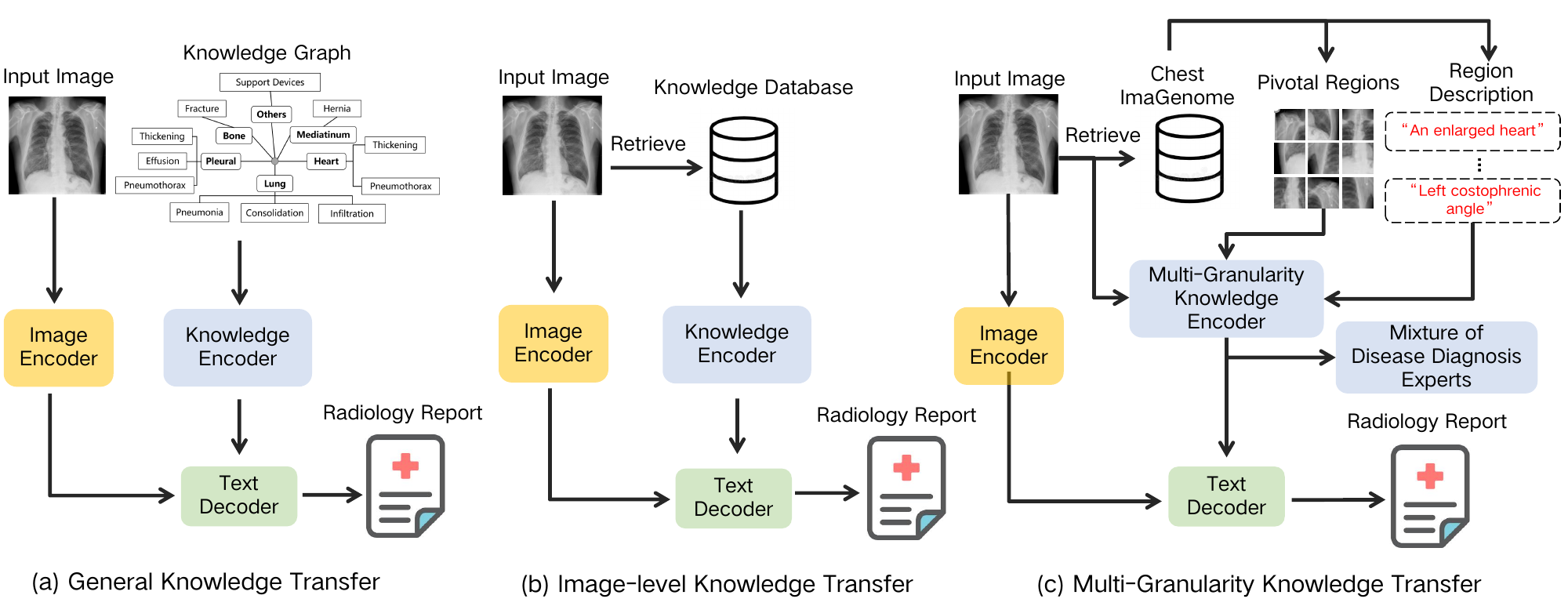}
\caption{Comparison of knowledge transfer approaches in radiology report generation. Existing methods achieve knowledge transfer through two primary approaches: (a) general knowledge transfer, which leverages symptom statistics to construct knowledge graph, and (b) image-level knowledge transfer, which  utilizes retrieved reports from knowledge database. 
(c) We propose a methodology that employs extra priors \cite{wu2021chest} to parse images and identify discriminative regions during training. This approach efficiently integrates global image context along with knowledge extracted from these critical areas by leveraging a multi-granularity knowledge encoder coupled with a mixture of disease diagnosis expert module.} \label{fig1}
\end{figure*}

Radiology report generation can automatically produce clinical sentences from X-ray images, thereby significantly enhancing the efficiency of radiologists. Compared to image captioning \cite{rotstein2024fusecap,wang2024mitigating,yao2025hifi} in natural scenes, radiology report generation is more challenging because it requires medical knowledge to accurately identify diseases and describe them in a professional manner. In order to leverage medical knowledge, existing methods \cite{bu2024instance,jin2024promptmrg,hu2022graph,huang2023kiut,liu2021exploring} have proposed different knowledge transfer approaches.
For example,
as shown in Fig.\ref{fig1}(a), existing methods \cite{hu2022graph,huang2023kiut} typically construct knowledge graph  and incorporate general medical expertise to enhance report generation. Moreover, some approaches \cite{bu2024instance,jin2024promptmrg,liu2021exploring} enhance the integration of image-level knowledge by utilizing retrieved expert reports from databases based on the input image.

However, these methods \cite{bu2024instance,jin2024promptmrg,hu2022graph,huang2023kiut,liu2021exploring} often overlook the critical role of medical knowledge in accurately capturing and describing  pivotal organ and lesion areas, which are essential for generating precise radiology reports. This poses a challenge because the varying size scales of discriminative regions require models to possess the capability to extract and aggregate knowledge embeddings at multiple levels of granularity. 
There have been recent attempts  \cite{chen2024fine,ref-7, zhou2025learnable} to employ a detection framework to select and fuse region feature.  This approach increases the model complexity and decrease the efficiency of knowledge transfer.

Accordingly, as shown in Fig.\ref{fig1}(c), we herein propose a novel and compact Multi-Granularity Expert Knowledge Transfer method to encode and aggregate  knowledge embeddings at multiple levels of granularity. First, we encode global knowledge embeddings using a pre-trained medical CLIP model \cite{you2023cxr}, which provides contextual medical knowledge. Moreover, we devise a novel Fine-Grained Knowledge Distillation (FGKD) training task which efficiently extract fine-grained knowledge. Specifically, the FGKD training task contains two types of embeddings:
teacher embeddings and student embeddings. 
Teacher embeddings are encoded by leveraging supplementary region bounding box and phrase description priors derived from Chest ImaGenome \cite{wu2021chest} or large foundation models \cite{wu2023medklip}.  Student embeddings are subsequently learned from the teacher embeddings through the process of knowledge distillation. During inference, the student embeddings are applied to augment fine-grained knowledge, whereas the teacher embeddings are discarded. Therefore, this approach incurs negligible computational costs and eliminates the need for additional priors. Finally, to further enhance multi-granularity knowledge transfer, we propose a novel classifier, namely Mixture of Disease Diagnosis Experts, which is built upon detailed text description of disease symptoms and the Mixture of Experts (MoE) approach \cite{masoudnia2014mixture}. 
In summary, the contributions of this paper are as follows:
\begin{itemize}
    \item
To the best of our knowledge, \textbf{EMGKT} is the first multi-granularity knowledge transfer approach  for radiology report generation, in which a novel fine-grained knowledge distillation training task is devised to achieve fine-grained knowledge encoding.
\end{itemize}
\begin{itemize}
    \item We propose a mixture of disease diagnosis experts classifier to promote knowledge transfer, which is built upon detailed text description of disease symptoms and the MoE approach.
\end{itemize}
\begin{itemize}
    \item Our \textbf{EMGKT} is compact and orthogonal to other radiology report generation networks. Experimental results show that \textbf{EMGKT} can be efficiently applied to various baselines and significantly improve report generation performance.
\end{itemize}

\section{Related Work}
\label{sec:RW}
\subsection{Image Caption Generation}

Image caption generation 
\cite{johnson2016densecap,wang2022ofa,li2023blip} has garnered significant attention within the field of computer vision. It offers crucial support for comprehending and interpreting the real world by automatically generating descriptive captions for images \cite{johnson2016densecap,wang2022ofa,li2023blip,yang2016review}. For example, Xu et al. \cite{xu2015show} presented an attention-based model capable of autonomously directing its focus on salient objects within images and generating descriptive content. Moreover, the applications of vision-language models \cite{chen2023vlp} have further promoted the development of image caption generation. For instance, Li et al. \cite{li2023blip} proposed to advance the general understanding of multi-modal data by pre-training a BLIP model on a vast corpus of image-text pairs. This approach enhances the model's ability to integrate and interpret information across visual and textual domains, consequently elevating the performance of image caption generation. \\
\subsection{Radiology Report Generation}

Compared to image caption generation, radiology report generation  is a less explored area, primarily due to its inherent complexities \cite{bu2024instance,hou2023organ,luo2024textual,chen2024fine,ref-7,ref-19,ref-20}. In response to these distinct challenges, existing approaches for radiology report generation \cite{bu2024instance,hou2023organ,luo2024textual,chen2024fine,ref-7,ref-19,ref-20}  can be classified into three categories. 
The first class of methods typically aims to address the misalignment issue between the medical vision and text modalities. 
For instance, Chen et al. \cite{ref-20} developed a shared memory framework designed to document the correspondence between image and text data, thereby enhancing cross-modal interaction and generation capabilities. Hou et al. \cite{hou2023organ} proposed to maintain consistency between images and lengthy reports by modeling the semantic structure of reports.
While Luo \cite{luo2024textual} et al. devised pseudo words to eliminate modality differences and construct a unified common feature space for images and texts. Zhou et al. \cite{zhou2025learnable} devised a fine-grained
visual-text alignment and fusion strategy that ensures consistency
across multi-source cross-attention maps for precise alignment.  
\begin{figure*}[t]
\includegraphics[width=\textwidth]{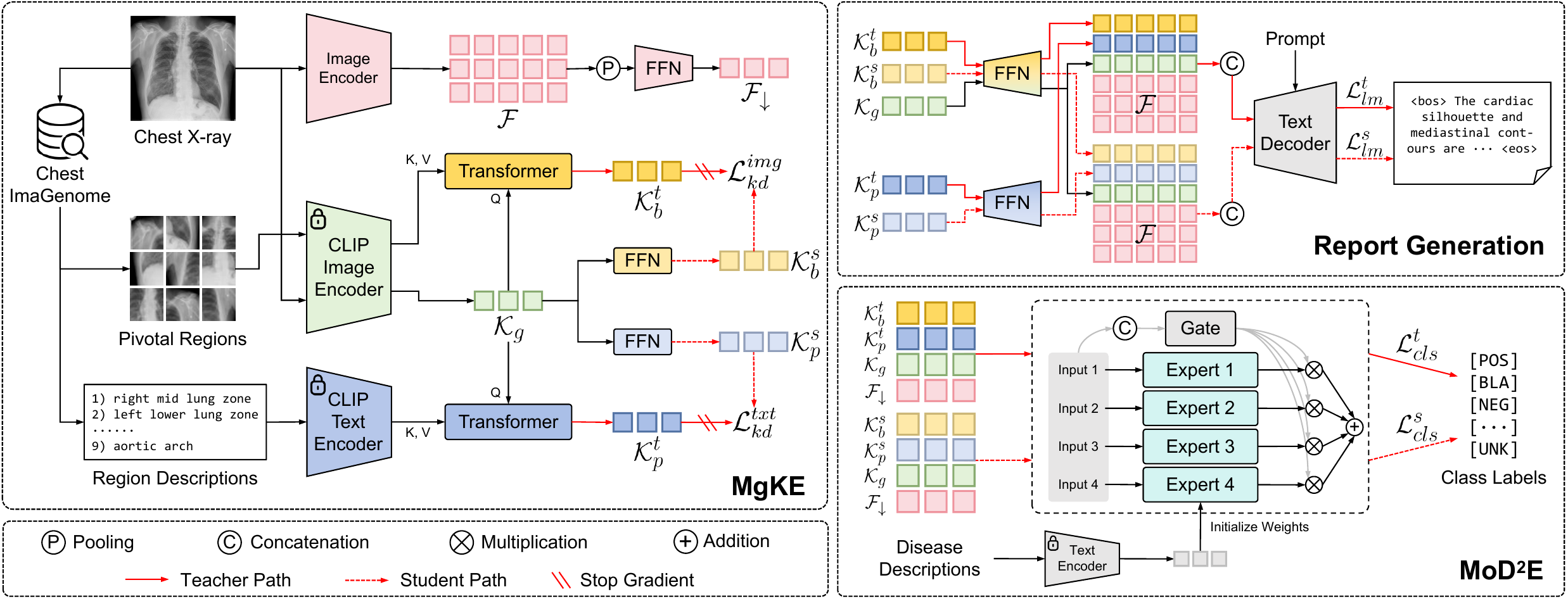}
\caption{Overview of our method \textbf{EMGKT}  during training phrase. Multi-granularity Knowledge Encoder (MgKE) efficiently encodes and integrates multi-granularity knowledge embeddings, which includes a novel fine-grained knowledge distillation approach. Mixture of Disease Diagnosis Experts (MoD$^{2}$E) further enhances disease knowledge by initializing classifiers using text embeddings of detailed disease descriptions; and subsequently fuses classification scores using a mixture of experts approach.} \label{overview}
\end{figure*}

The second category of methods tend to focus on extracting fine-grained features for important region description. For example, Tim et al. \cite{ref-7} directly extracted local visual features of anatomical regions based on off-the-shelf detection frameworks \cite{NIPS2015_14bfa6bb}. Moreover, Chen et al. \cite{chen2024fine} proposed an adaptive detection framework to obtain anomalous regions. However, the implementation of these detection frameworks tends to escalate model complexity and ignore the efficiency of knowledge transfer processes. Accordingly, we herein introduce a novel  knowledge distillation training task to achieve efficient encoding of fine-grained knowledge.
The third class of methods typically incorporate prior medical knowledge to generate clinical sentences \cite{bu2024instance,huang2023kiut}. For example, Huang et al. \cite{huang2023kiut} proposed to incorporate clinical knowledge by construct symptom graphs during the decoding stage. Jin et al. \cite{jin2024promptmrg} proposed to integrate disease diagnosis knowledge by converting the diagnostic results into token prompts that explicitly direct the generation process. While Bu et al. \cite{bu2024instance} built image-level expert knowledge for each disease, providing expert insights during the decoding process. However, these methods overlook the significance of incorporating multi-granularity knowledge in the generation of medical reports. Accordingly, we herein propose a novel method that enables exisiting methods to efficiently learn and aggreate multi-granularity knowledge. In the experimentation section, we further demonstrate that our approach can be  applied to various radiology report generation methods.

\subsection{Knowledge Distillation}

Knowledge distillation (KD) \cite{hinton2015distilling} is a widely adopted technique for transferring knowledge from a large, well-trained teacher model to a lightweight student model, thereby improving the student's performance without incurring additional inference cost. Existing KD approaches can be broadly divided into three categories. The first category, response-based distillation \cite{hinton2015distilling,zhao2022decoupled}, encourages the student to mimic the teacher's output logits or soft predictions. For instance, Hinton et al. \cite{hinton2015distilling} introduced the concept of ``soft targets'' to convey the teacher's dark knowledge, while Zhao et al. \cite{zhao2022decoupled} decoupled the classical KD loss into target-class and non-target-class components to improve distillation efficiency. The second category, feature-based distillation \cite{romero2014fitnets,heo2019comprehensive}, instead aligns the intermediate feature representations between the teacher and student networks. Romero et al. \cite{romero2014fitnets} proposed FitNets to guide the student using the teacher's hidden layer activations, and Heo et al. \cite{heo2019comprehensive} further introduced margin-based activation transfer to preserve more informative features. The third category, relation-based distillation \cite{park2019relational,tung2019similarity}, focuses on transferring structural knowledge, such as the relationships or similarities among multiple samples or feature dimensions, rather than the individual outputs. Park et al. \cite{park2019relational} proposed relational knowledge distillation to capture mutual relations of data examples, while Tung et al. \cite{tung2019similarity} preserved pairwise activation similarities across the teacher and student networks. However, these methods are typically designed for generic classification or recognition tasks and overlook the fine-grained, region-level knowledge that is essential for medical imaging, where subtle anatomical and pathological details often determine diagnostic accuracy. Accordingly, we herein propose a fine-grained knowledge distillation approach tailored to radiology report generation, which efficiently transfers multi-granularity knowledge to guide the encoding of both global and local disease-related representations.
\begin{figure*}[t]
\centering
\includegraphics[width=1.\textwidth]{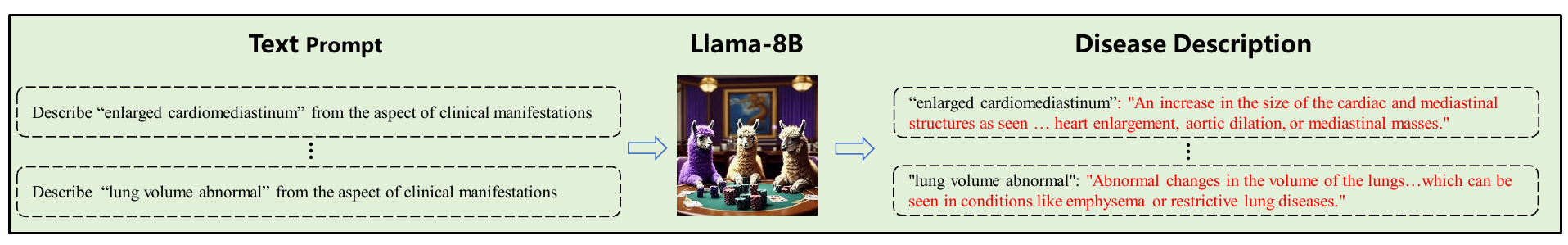}
\caption{Illustration of disease description generation using LLM.} \label{fig_Moe}
\end{figure*}

\section{Method}
\textbf{EMGKT}  can be efficiently applied to existing radiology report generation methods \cite{jin2024promptmrg,ref-20}. In this section, we take one representative work PromptMRG \cite{jin2024promptmrg} as an example. As illustrated in Fig.\ref{overview}, our baseline framework comprises two main stages: feature encoding and report generation. During the encoding stage, each input image $\mathcal{I}$ is processed by an image encoder \cite{ref-26} and subsequently transformed into image features $\mathcal{F} \in \mathbb{R}^{B\times L\times D}$, where $B$ is the batch size, $L$ is the number of the processed patches and $D$ is the dimension. In the  report generation stage, image features $\mathcal{F}$ and the text embeddings $\mathcal{T}_{n-1}$ of previously generated word $\mathcal{W}_{n-1}$ are fed into the text decoder~\cite{ref-29}, which then produces the log probability $\mathcal{P}_{n}$ of the next word.
\begin{equation}
\mathcal{P}_{n}=f_{d}(\mathcal{F},\mathcal{T}_{n-1}), \quad \mathcal{F} =f_{i}(\mathcal{I}), \quad \text{and} \quad \mathcal{T}_{n-1} =f_{t}(\mathcal{W}_{n-1}),
\end{equation}
where $f_{d}$, $f_{i}$, and $f_{t}$ denote text decoder, image encoder, and text encoder, respectively. Finally, the $n$-th word can be predicted after applying the softmax function to $\mathcal{P}_{n}$, as formulated below:
\begin{equation}
\mathcal{W}_{n}=\text{softmax}(\mathcal{P}_{n}).
\end{equation}

As shown in Fig.\ref{overview}, \textbf{EMGKT} consists of two main modules, namely, Multi-granularity Knowledge Encoder (MgKE) (in Section \ref{sec:MgKE}) and Mixture of Disease Diagnosis Experts (MoD$^{2}$E) (in Section \ref{sec:MoDE}).  

\subsection{Multi-Granularity Knowledge Encoder}
\label{sec:MgKE}
Precise descriptions of pivotal regions are essential for radiology reports. 
In order to capture different sizes of pivotal regions and promote precise region descriptions, we propose a novel Multi-granularity Knowledge Encoder (MgKE) module. MgKE constructs and integrates two types of knowledge embeddings: global $\mathcal{K}_{g}$ and fine-grained region $\mathcal{K}_{f}$ ; each type of embeddings represents one granularity-specific, which are extracted from a medical CLIP model \cite{you2023cxr}. 

\noindent \textbf{Global Knowledge Embedding}. As shown in Fig.\ref{overview}, given a X-ray Image $\mathcal{I} \in \mathbb{R}^{H\times W\times 3}$, we extract global knowledge embedding $\mathcal{K}_{g}$ using a medical CLIP image encoder \cite{you2023cxr}. The CLIP model is pre-trained using image-report pairs, thereby providing medical context knowledge. This process can be represented as:
\begin{equation}
\label{eq_kg}
\mathcal{K}_{g}=f_{c}(\mathcal{I}),
\end{equation}
where $f_{c}$ denotes the CLIP image encoder.

\noindent \textbf{Fine-Grained Knowledge Embedding}. The encoding of fine-grained knowledge often necessitates comprehensive priors \cite{wu2021chest,wu2023medklip}, posing challenges to the efficient extraction of fine-grained knowledge embeddings.
To address this issue, we devise a knowledge distillation task that introduces negligible computation during inference. 

First, as shown in Fig.\ref{overview}, for each image $\mathcal{I}$, we obtain a set of region  priors of bounding boxes $B=\{b_{i}\}^{M}_{i=1}$ and text descriptions $P=\{p_{i}\}^{M}_{i=1}$ from Chest ImaGenome  \cite{wu2021chest}. These priors can also be obtained using large foundation models following \cite{zhou2025learnable}.
Subsequently, we utilize $B=\{b_{i}\}^{M}_{i=1}$ and $P=\{p_{i}\}^{M}_{i=1}$ to extract teacher knowledge embeddings, which can be formulated  as follows:
\begin{equation}
\mathcal{K}^{t}_{b}= \Phi_{b}(f_{c}(\mathcal{I},B)) \quad  \text{and}  \quad \mathcal{K}^{t}_{p}=  \Phi_{p}(f_{p}(P)),
\end{equation}
where $\mathcal{K}^{t}_{b}$ and $\mathcal{K}^{t}_{p}$
denotes teacher vision embeddings and teacher text embeddings, respectively; while $\Phi_{b}(\cdot)$ and $\Phi_{p}(\cdot)$ represent transformer attention modules \cite{ref-31} and $f_{p}(\cdot)$ denotes the CLIP text encoder.

Subsequently, we construct a teacher feature stream by augmenting image feature $\mathcal{F}$ with these teacher embeddings as following:
\begin{equation}
\mathcal{P}^{t}_{n}=f_{d}(\mathcal{F}^{t},\mathcal{T}_{n-1}),
\end{equation}
\begin{equation}
\quad \mathcal{F}^{t} = Concat(\mathcal{F}, FFN(\mathcal{K}_{g}), FFN(\mathcal{K}^{t}_{b}), FFN(\mathcal{K}^{t}_{p})),
\end{equation}
where $Concat(\cdot)$ represents feature concatenation and $FFN(\cdot)$ denotes feed-forward network \cite{ref-31} to align the dimensions of knowledge embeddings and image features.
As the teacher embeddings $\mathcal{K}^{t}_{b}$ and $\mathcal{K}^{t}_{p}$ are obtained using extra labels which increases the computation cost during inference, we creatively propose to extract fine-grained knowledge $\mathcal{K}_{f}$ from global knowledge $\mathcal{K}_{g}$ using $\mathcal{K}^{t}_{b}$ and $\mathcal{K}^{t}_{p}$ as distillation supervision signal:

\begin{equation}
\mathcal{K}^{s}_{b}= FFN(\mathcal{K}_{g}) \quad \text{and} \quad \mathcal{K}^{s}_{p}= FFN(\mathcal{K}_{g}) 
\end{equation}
\begin{equation}
\mathcal{L}_{kd} = ||\mathcal{K}^{s}_{b}-sg(\mathcal{K}^{t}_{b})||^{2} +  ||\mathcal{K}^{s}_{p}-sg(\mathcal{K}^{t}_{p})||^{2},
\end{equation}
where $||\cdot||^{2}$ represents $L_{2}$ loss function; $sg(\cdot)$ stands for the stop-gradient operator. 

Finally, we use the student feature stream for report generation as follows:
\begin{equation}
\mathcal{P}^{s}_{n}=f_{d}(\mathcal{F}^{s},\mathcal{T}_{n-1}),
\end{equation}
\begin{equation}
   \mathcal{F}^{s} = Concat(\mathcal{F}, FFN(\mathcal{K}_{g}),  FFN(\mathcal{K}^{s}_{b}), FFN(\mathcal{K}^{s}_{p})).
\end{equation}
Notably, the teacher embeddings and student embeddings are trained using the same loss functions (see details in Section \ref{sec:TO}). Excluding the parameters of the CLIP image encoder, all other parameters are iteratively updated using the overall loss functions.  Moreover, teacher embedding $\mathcal{K}^{t}_{b}$ and $\mathcal{K}^{t}_{p}$ are only constructed for training while abandoned during inference. We present the overall framework of our method during inference stage in Fig. \ref{overview2}.

\subsection{Mixture of Disease Diagnosis Experts}
\label{sec:MoDE}
Disease classification is typically used as an auxiliary task for improving the diagnostic accuracy of radiology report generation \cite{ref-20,ref-33,jin2024promptmrg}. Existing methods typically construct disease classifiers that are randomly initialized and optimized using image-based labels. However, compared to common object classification \cite{chen2021review,rao2021global}, disease identification pays more attention to the complexity and professionalism of the medical field, requiring medical expert knowledge \cite{ref-33,jin2024promptmrg,kim2022transfer}.

To address the above issues, we introduce a novel classifier,termed as Mixture of Disease Diagnosis Experts classifier (MoD$^{2}$E), designed to enhance the extraction of multi-granularity knowledge. As shown in Fig. \ref{fig_Moe}, we firstly devise medical prompts to generate detailed description from a large language model \cite{touvron2023llama} for each disease. As illustrated in Fig.\ref{overview}, the disease descriptions are encoded into text embeddings which are then used to initialize disease classifiers \cite{liao2022gen}. 
Moreover, we model different classifiers as distinct diagnosis experts for varying granularity and modality features, and fuse their classification scores using the MoE approach \cite{masoudnia2014mixture}. This process can be expressed as:
\begin{equation}
\mathcal{C} =\sum_{i=1}^{M} \mathcal{G}_{i}(\mathbf{x}; \boldsymbol{\Theta}_{i}) f_{i}\left(\mathbf{x}_i ; \mathbf{W}_{i}\right),
\end{equation}
\begin{equation}
\mathcal{G}_{i}(\mathbf{x}; \boldsymbol{\Theta}_{i})=\operatorname{softmax}(g(\mathbf{x} ; \boldsymbol{\Theta}_{i})), \end{equation} 
\begin{equation}
 \quad x_i  \in \{\mathcal{F_{\downarrow}}, \mathcal{K}_{g},  \mathcal{K}_{b}, \mathcal{K}_{p}\}, \text{and} \quad \mathcal{F_{\downarrow}} = FFN(Avg(\mathcal{F}))
\end{equation}
where $g(\mathbf{x} ; \boldsymbol{\Theta}_{i})$ and $f_{i}\left(\mathbf{x}_i ; \mathbf{W}_{i}\right)$ represent weight assignment and disease classification functions \cite{masoudnia2014mixture}, respectively. $Avg(\cdot)$ denotes the average pooling operation. Accordingly, the disease classification loss for the teacher embeddings scores $\mathcal{C}^{t}$  can be denoted as follows:
\begin{equation}
\mathcal{C}^{t} =\sum_{i=1}^{M} \mathcal{G}_{i}(\mathcal{F_{\downarrow}}, \mathcal{K}_{g}, \mathcal{K}^{t}_{b}, \mathcal{K}^{t}_{p} ; \boldsymbol{\Theta}_{i}) f_{i}\left(\mathbf{x}_{i}; \mathbf{W}_{i}\right),
\end{equation}
While the disease classification scores for the student embeddings stream $\mathcal{C}^{s}$  can be represented as follows:
\begin{equation}
\mathcal{C}^{s} =\sum_{i=1}^{M} \mathcal{G}_{i}(\mathcal{F_{\downarrow}}, \mathcal{K}_{g}, \mathcal{K}^{s}_{b}, \mathcal{K}^{s}_{p}; \boldsymbol{\Theta_{i}}) f_{i}\left(\mathbf{x}_{i}; \mathbf{W}_{i}\right),
\end{equation}
\begin{table*}[t]
\caption{Comparison between baselines and the improved network with \textbf{EMGKT}. $\triangle$ denotes the improvements compared to the baselines. * denotes our re-implementation of baselines.}\label{tab1}
\centering
{\scriptsize
\begin{tabular}{l c c c c c c}
\toprule
 \multirow{2}{*}{Method} & \multicolumn{6}{c}{NLG Metrics} \\ \cmidrule(lr){2-7}
   & BLEU-1 & BLEU-2& BLEU-3& BLEU-4 & METEOR & ROUGE-L\\
\midrule
\multicolumn{7}{c}{Experimental results on MIMIC-CXR dataset.} \\
\midrule
R2GenCMN\textsuperscript{*}~\cite{ref-20} & 0.344 &0.210 &0.139 &0.098 &0.136&0.275\\
  \textbf{+Ours} &  \textbf{0.356} & \textbf{0.219}& \textbf{0.146} & \textbf{0.105}& \textbf{0.139}
& \textbf{0.278} \\
  $\triangle$  
 & \textbf{0.012} &\textbf{0.009} &\textbf{0.007} &\textbf{0.007}&\textbf{0.003}
&\textbf{0.003}\\
\midrule
PromptMRG\textsuperscript{*}~\cite{jin2024promptmrg} & 0.395&0.236&0.154 &0.112  &0.154  &0.267\\
  \textbf{+Ours} &  \textbf{0.409} & \textbf{0.250}& \textbf{0.166} & \textbf{0.117}& \textbf{0.157}
& \textbf{0.273} \\
  $\triangle$  
 & \textbf{0.014} &\textbf{0.014} &\textbf{0.012} &\textbf{0.005}&\textbf{0.003}
&\textbf{0.006}\\
\midrule
\multicolumn{7}{c}{Experimental Results on IU-XRAY dataset.} \\
\midrule
R2GenCMN\textsuperscript{*}~\cite{ref-20} &0.397 &0.230&0.138&0.085&0.140 &0.281 \\ 
   \textbf{+Ours} &  \textbf{0.405} & \textbf{0.235}&  \textbf{0.140} & \textbf{0.088} & \textbf{0.143}& \textbf{0.283}\\
$\triangle$  
& \textbf{0.008} &\textbf{0.005} &\textbf{0.002} &\textbf{0.003}&\textbf{0.003}
&\textbf{0.002}\\
\midrule
PromptMRG\textsuperscript{*}~\cite{jin2024promptmrg} & 0.408 &0.241&0.152 &0.101  &0.162  &0.310 \\ 
 \textbf{+Ours} & \textbf{0.428} &\textbf{0.261}
  &\textbf{0.168} &\textbf{0.113}
  &\textbf{0.167}
  &\textbf{0.321}\\
  $\triangle$  
 & \textbf{0.020} &\textbf{0.020} &\textbf{0.016} &\textbf{0.012}&\textbf{0.005}
&\textbf{0.011}\\
\bottomrule
\end{tabular}
}
\end{table*}
\subsection{Training Objective}
\label{sec:TO}
Following previous works \cite{ref-20,ref-33,jin2024promptmrg}, we utilize language modeling loss $\mathcal{L}_{lm}$ to optimize the report generation task. The formulation of $\mathcal{L}_{lm}$ is represented as follows:

\begin{equation}
\mathcal{L}_{lm}= \mathcal{L}^{t}_{lm}+ \mathcal{L}^{s}_{lm},\label{eq:lm}
\end{equation}

\begin{equation}
\mathcal{L}^{t}_{lm}=-\sum_{n=1}^N \log p\left(T_n \mid T_1, \ldots, T_{n-1}, \mathcal{F},\mathcal{K}_{g}, \mathcal{K}^{t}_{b},\mathcal{K}^{t}_{p}\right),\label{eq:lm_t}
\end{equation}

\begin{equation}
\mathcal{L}^{s}_{lm}=-\sum_{n=1}^N \log p\left(T_n \mid T_1, \ldots, T_{n-1}, \mathcal{F},\mathcal{K}_{g}, \mathcal{K}^{s}_{b}, \mathcal{K}^{s}_{p}\right),\label{eq:lm_s}
\end{equation}
where $\mathcal{L}^{t}_{lm}$ and $\mathcal{L}^{s}_{lm}$ denotes language modeling loss for feature streams of teacher and student knowledge embeddings, respectively.

Moreover, we employ Cross-Entropy loss $\mathcal{L}_{ce}$ to supervise the disease classification task, which can be denoted as follows:
\begin{equation}
\mathcal{L}_{cls}=\mathcal{L}_{ce}(\mathcal{C}^{t}, \hat{\mathcal{C}}) + \mathcal{L}_{ce}(\mathcal{C}^{s}, \hat{\mathcal{C}}),
\end{equation}
where $\mathcal{C}^{t}$ and $\mathcal{C}^{s}
$ represent the predicted disease logit probabilities for  feature streams of teacher and student knowledge embeddings, respectively; while $\hat{\mathcal{C}}$ denotes ground-truth disease labels.

Finally, the overall loss function can  be formulated as follows:
\begin{equation}
\mathcal{L}_{all}=\alpha\mathcal{L}_{lm} +\beta\mathcal{L}_{cls} + \gamma \mathcal{L}_{kd} .
\end{equation}
where $\alpha$ is set to 1.0 following that in \cite{jin2024promptmrg}; while $\gamma$  and $\beta$ are empirically set to 10.0 and 8.0, respectively.

\section{Experiment}
\subsection{Dataset and Evaluation Metrics}

\textbf{Dataset.} IU X-ray\footnote{https://openi.nlm.nih.gov/.}~\cite{ref-35} and MIMIC-CXR\footnote{https://physionet.org/content/mimic-cxr/2.0.0/.}~\cite{ref-36}  are the two most popular benchmarks for radiology report generation. 
IU X-ray is a small dataset, containing 7,470 chest X-ray images and 3,955 corresponding reports; while MIMIC-CXR is a large-scale dataset,including 377,110 images and 227,835 corresponding reports. To ensure consistency and fairness in the comparative analysis, we adopted the data processing methodologies employed by the baseline models~\cite{ref-18,ref-20,jin2024promptmrg}.

\textbf{Evaluation Metrics.} Following previous works\cite{ref-20,ref-33,jin2024promptmrg}, we evaluate our method on natural language generation (NLG) metrics including BLEU~\cite{ref-38}, METEOR~\cite{ref-39} and ROUGE-L~\cite{ref-40}, which are widely used to assess the fluency and accuracy of generated reports. Moreover, we follow \cite{jin2024promptmrg} and use the model trained on MIMIC-CXR training set to directly perform the evaluation on the whole set of IU X-Ray. We also employ the CE metrics for disease classifications, including precision, recall, and F1-score.
\cite{jin2024promptmrg}. 
\begin{table}[t]
    \centering
    \caption{The comparison of the clinical efficacy metrics on
MIMIC-CXR dataset.}
    \label{tab:ce}
    \begin{tabular}{@{}lccc@{}}
        \toprule
        \textbf{Method} & \textbf{Precision} & \textbf{Recall} & \textbf{F1-Score} \\
        \midrule
        R2GenCMN \cite{ref-20} & 0.334 & 0.275 & 0.278 \\
        GSKET \cite{yang2022knowledge} & 0.458 &0.348 & 0.371 \\
        Clinical-BERT \cite{yan2022clinical} & 0.397 & 0.435 & 0.415 \\
        KiUT \cite{huang2023kiut} &0.371 & 0.318 & 0.321 \\
        DCL \cite{li2023dynamic} & 0.471 & 0.352 & 0.373 \\
        METransformer \cite{wang2023metransformer} & 0.364 & 0.309 & 0.311 \\
        PromptMRG  \cite{jin2024promptmrg} & 0.501 & 0.509 & 0.476 \\
         \hline
         R2GenCMN \cite{ref-20} + Ours  & \textbf{0.372} & \textbf{0.289} & \textbf{0.306} \\
         PromptMRG \cite{jin2024promptmrg}  + Ours  & \textbf{0.517} & \textbf{0.518} & \textbf{0.488} \\
        \bottomrule
    \end{tabular}
\end{table}

\begin{figure*}[t]
\includegraphics[width=\textwidth]{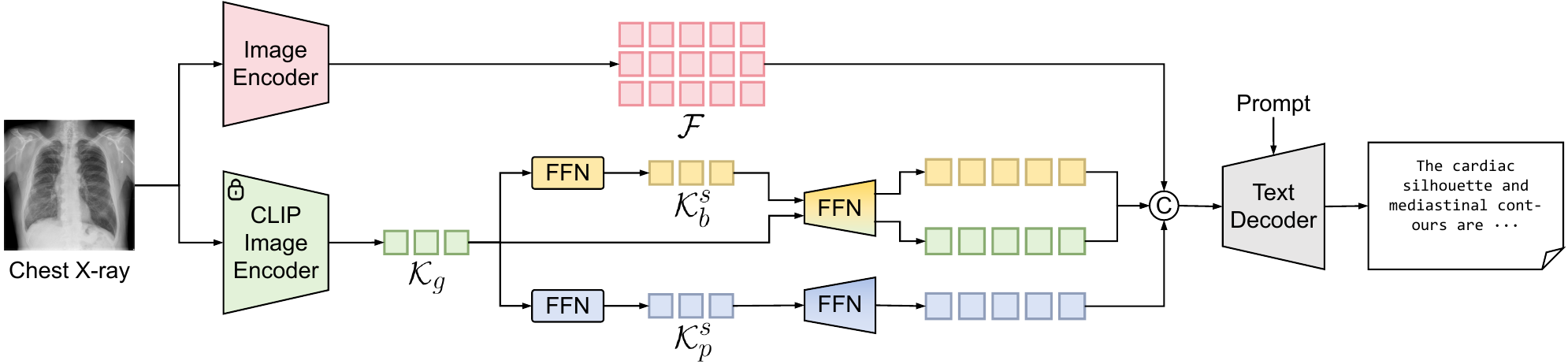}
\caption{Overview of \textbf{EMGKT} during inference phrase.} \label{overview2}
\end{figure*}

\begin{table*}[h]
\centering
\caption{Ablation experiments of efficient multi-granularity knowledge transfer.}\label{tab2}
{\scriptsize
\resizebox{0.98\textwidth}{!}{
\begin{tabular}{c c c c c c c c c}
\toprule
\begin{tabular}[c]{@{}c@{}} Multi-Granularity  \\ Knowledge Encoder \end{tabular}&\begin{tabular}[c]{@{}c@{}}Mixture of Disease \\  Diagnosis Experts \end{tabular}   &  BLEU-1 & BLEU-2 & BLEU-3 & BLEU-4 & METEOR & ROUGE-L & \begin{tabular} [c]{@{}c@{}} Inference \\ ms/img \end{tabular}\\
\midrule
 &  & 0.395&0.236&0.154 &0.112  &0.154  &0.267&124\\
\checkmark&  & 0.405 &0.246
&0.162& 0.114 &0.154&0.272 &137\\  
\midrule
 \checkmark& \checkmark&   \textbf{0.409} & \textbf{0.250}& \textbf{0.166} & \textbf{0.117}& \textbf{0.157}
& \textbf{0.273}  &137\\
\bottomrule
\end{tabular}
}}
\end{table*}
\begin{table*}[h]
\centering
\caption{Ablation experiments on the structure of multi-granularity knowledge encoder.}\label{tab3}
{\scriptsize
\resizebox{1.0\textwidth}{!}{
\begin{tabular}{ccccccccc}
\toprule
\begin{tabular}[c]{@{}c@{}}Global \\ Embeddings\end{tabular}&\begin{tabular}[c]{@{}c@{}}Fine-Grained \\ Vision Embeddings\end{tabular} &\begin{tabular}[c]{@{}c@{}}Fine-Grained \\ Text Embeddings\end{tabular}   &  BLEU-1 & BLEU-2 & BLEU-3 & BLEU-4 & METEOR & ROUGE-L\\
\midrule
&  & &    0.395&0.236&0.154 &0.112  &0.154  &0.267\\
$\checkmark$&  & &    0.401&0.243&  0.160  & 0.112&0.156&0.271\\
$\checkmark$&$\checkmark$ &&  0.406 &0.247
&0.162& 0.114 &0.154&0.272\\
\midrule
$\checkmark$&$\checkmark$ &$\checkmark$  &\textbf{0.409}  &\textbf{0.250}& \textbf{0.166} & \textbf{0.117}& \textbf{0.157} & \textbf{0.273}  \\
\bottomrule
\end{tabular}
}
}
\end{table*}

\subsection{Experiments Results and Analyses}
\textbf{Comparison with Baselines.} To demonstrate the effectiveness and transferability of \textbf{EMGKT}, we evaluate it on three representative baseline models: R2GenCMN~\cite{ref-20} and PromptMRG~\cite{jin2024promptmrg}. We integrate \textbf{EMGKT} into each of these baselines and compare the enhanced models against their original counterparts, with the results summarized in Table~\ref{tab1}.

It can be observed that \textbf{EMGKT} improves all NLG metrics across the three baselines, indicating that it can effectively integrate multi-granularity knowledge to generate more accurate reports. We further conduct cross-dataset knowledge transfer experiments on the IU X-Ray dataset, where models trained on the MIMIC-CXR training set are directly evaluated on IU X-Ray without any additional fine-tuning. As shown in Table~\ref{tab1}, even in the absence of region-location or region-description annotations for IU X-Ray, \textbf{EMGKT} consistently improves report generation performance by effectively transferring the knowledge learned from MIMIC-CXR. Moreover, as illustrated in Table~\ref{tab:ce}, our approach substantially improves the disease classification performance of all baseline models, outperforming several strong existing methods. The overall framework of \textbf{EMGKT} during the inference stage is presented in Fig.~\ref{overview2}.

\begin{table*}[t]
\centering
\caption{Ablation experiments on the structure of disease classifiers.}\label{tab4}
{\scriptsize
\resizebox{0.8\textwidth}{!}{
\begin{tabular}{c c c c c c c c }
\toprule
\begin{tabular}[c]{@{}c@{}} CLIP Embedding \\Initialization  \end{tabular}&\begin{tabular}[c]{@{}c@{}} Mixture of \\Experts  \end{tabular}  &  BLEU-1 & BLEU-2 & BLEU-3 & BLEU-4 & METEOR & ROUGE-L \\
\midrule
 &  & 0.395&0.236&0.154 &0.112  &0.154  &0.267\\
\checkmark&  &  0.400  &  0.245
  &    0.161  & 0.113 &  0.152  &  0.272 \\  
& \checkmark&  0.402  &  0.243
  &    0.158  & 0.110 &  0.155 &  0.270 \\
\midrule
 \checkmark& \checkmark&   \textbf{0.409} & \textbf{0.250}& \textbf{0.166} & \textbf{0.117}& \textbf{0.157}
& \textbf{0.273}  \\
\bottomrule
\end{tabular}
}}
\end{table*}

\begin{figure*}[h]
\includegraphics[width=\textwidth]{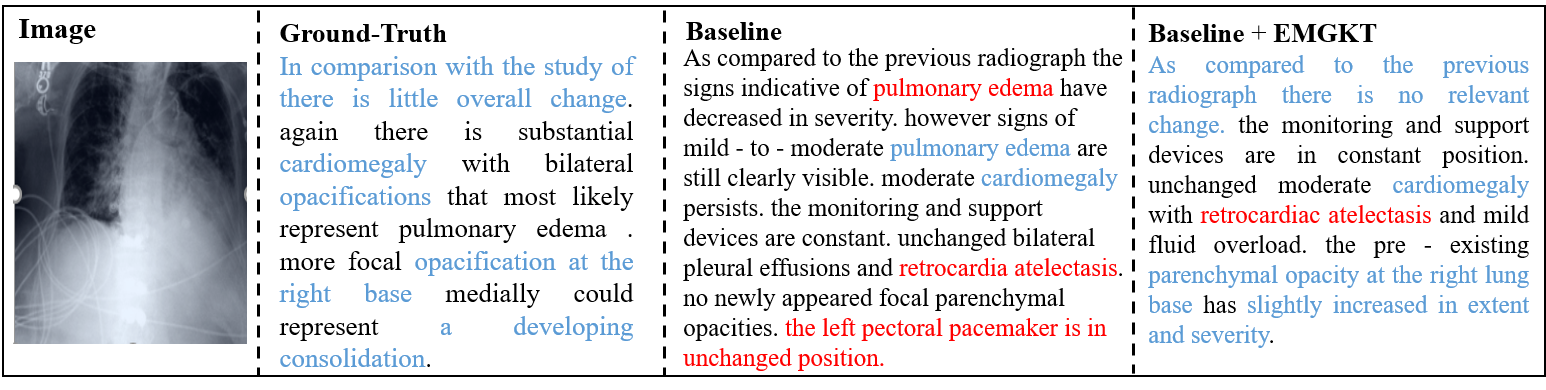}
\caption{Visualization results of the baseline and our method. Content rendered in blue font signifies alignment with the ground-truth, whereas content in red font denotes discrepancies or inaccuracies.}
\label{visual}
\end{figure*}

\textbf{Ablation Study.} To verify the effectiveness of each component of \textbf{EMGKT}, we conducted various ablation experiments using  PromptMRG~\cite{jin2024promptmrg} as baseline model on MIMIC-CXR test set. Experiment results are presented in Table \ref{tab2}$-$\ref{tab4}.

First, as shown in Table~\ref{tab2}, 
equipped with MgKE, the performance of the baseline is notably improved by 1.0\% in terms of BLEU-1 metric. Furthermore, 
when MoD$^{2}$E is integrated, the performance of the baseline improves by 0.50\% in terms of the BLEU-1 metric. When both MgKE and MoD$^{2}$E are employed, the performance of the baseline is consistently enhanced by 1.40\% in terms of the BLEU-1 metric. It can also be observed both MgKE and MoD$^{2}$E  introduce negligible computational overhead during inference, highlighting the knowledge transfer efficiency of \textbf{EMGKT}.

Secondly, we conducted experiments to demonstrate the architecture of MgKE, as illustrated in Table \ref{tab3}. Upon integrating global knowledge embeddings ($\mathcal{K}_{g}$ in Eq.(\ref{eq_kg})), the baseline performance across all report generation metrics is enhanced. Moreover, the incorporation of both fine-grained visual and textual embeddings further improves these metrics.
report generation metrics. 

Finally, we conduct ablation studies to verify the structure of MoD$^{2}$E, as shown in Table~\ref{tab4}. The CLIP embeddings are encoded using detailed disease descriptions, aiming to promote the disease knowledge extraction. It can be observed that this technique effectively improves  report generation metrics. Furthermore, the MoE approach can further improve all the report generation metrics. 

\section{Qualitative Analysis}
As shown in Fig.\ref{visual}, the baseline model exhibits suboptimal performance, primarily due to the incorrect identification and presentation of important regions. In contrast, our \textbf{EMGKT} significantly improves report generation by precisely pinpointing pivotal regions and refining the corresponding descriptions in the baseline model.

\section{Conclusion}
In this paper, we propose a novel Efficient Multi-Granularity Knowledge Transfer (\textbf{EMGKT}) approach for radiology report generation. \textbf{EMGKT} consists two main components, namely, Multi-granularity Knowledge Encoder (MgKE) and Mixture of Disease Diagnosis Experts (MoD$^{2}$E). 
MgKE extracts multi-granularity knowledge embeddings based on the CLIP model and a novel knowledge distillation task. While MoD$^{2}$E further promotes knowledge extraction by using text embeddings as the initialization parameters of classifiers and employing the MoE approach to handle different granularity features.
Finally, extensive experiments are conducted to demonstrate the effectiveness  and transferability of \textbf{EMGKT} on two widely-used benchmarks and various baselines.

\end{document}